\documentclass[sigconf,natbib=false]{acmart}
\usepackage[bibstyle=ACM-Reference-Format,backend=bibtex,maxnames=2]{biblatex}
\usepackage{comment}

\AtBeginDocument{
  \providecommand\BibTeX{{
    \normalfont B\kern-0.5em{\scshape i\kern-0.25em b}\kern-0.8em\TeX}}}

\copyrightyear{2022}
\acmYear{2022}
\setcopyright{acmcopyright}\acmConference[KDD '22]{Proceedings of the 28th ACM
SIGKDD Conference on Knowledge Discovery and Data Mining}{August 14--18,
2022}{Washington, DC, USA}
\acmBooktitle{Proceedings of the 28th ACM SIGKDD Conference on Knowledge
Discovery and Data Mining (KDD '22), August 14--18, 2022, Washington, DC, USA}
\acmPrice{15.00}
\acmDOI{10.1145/3534678.3539339}
\acmISBN{978-1-4503-9385-0/22/08}

\acmSubmissionID{rtfp1155}

\usepackage{multirow}
\usepackage{stmaryrd}
\usepackage{subcaption}
\usepackage{color, colortbl}
\usepackage{bbding}

\newcommand{\cmark}{\scriptsize{\Checkmark}}
\newcommand{\xmark}{\scriptsize{\XSolidBrush}}

\definecolor{grayshade}{gray}{0.9}
\newcommand{\advcell}[1]{
\cellcolor{grayshade}\textcolor{black}{#1}}

\begin{document}

\title{Local Evaluation of Time Series Anomaly Detection Algorithms}

\author{Alexis Huet}
\affiliation{
  \institution{Huawei Technologies Co., Ltd.}
  \country{France}}
\email{alexis.huet@huawei.com}

\author{Jose Manuel Navarro}
\affiliation{
  \institution{Huawei Technologies Co., Ltd.}
  \country{France}}
\email{jose.manuel.navarro@huawei.com}

\author{Dario Rossi}
\affiliation{
  \institution{Huawei Technologies Co., Ltd.}
  \country{France}}
\email{dario.rossi@huawei.com}

\renewcommand{\shortauthors}{Alexis Huet, Jose Manuel Navarro, \& Dario Rossi}

\begin{abstract}
In recent years, specific evaluation metrics for time series anomaly detection algorithms have been developed to handle the limitations of the classical precision and recall. However, such metrics are heuristically built as an aggregate of multiple desirable aspects, introduce parameters and wipe out the interpretability of the output. In this article, we first highlight the limitations of the classical precision/recall, as well as the main issues of the recent event-based metrics -- for instance, we show that an adversary algorithm can reach high precision and recall on almost any dataset under weak assumption. 
To cope with the above problems, we propose a theoretically grounded, robust, parameter-free and interpretable extension to precision/recall  metrics, based on the concept of ``affiliation'' between the ground truth and the prediction sets. Our metrics leverage measures of duration between ground truth and predictions, and have thus an intuitive interpretation. By further comparison against random sampling, we obtain a normalized precision/recall, quantifying how much a given set of results is better than a random baseline prediction.
By construction, our approach keeps the evaluation local regarding ground truth events, enabling fine-grained visualization and interpretation of algorithmic results.
We compare our proposal against various public time series anomaly detection datasets, algorithms and metrics. We further derive theoretical properties of the affiliation metrics that give explicit expectations about their behavior and ensure robustness against adversary strategies. 
\end{abstract} 



\maketitle

\section{Introduction}

Time series anomaly detection is the field consisting in detecting elements of a time series that behave differently from the rest of the data.
This field attracted interest in recent years with the rise of monitoring systems collecting a large amount of data over time, mainly for the purpose of troubleshooting and security. 
Many scientific domains are involved: water control industrial systems~\cite{dataset:swat--goh2016dataset, dataset:water-hai--shin2020hai}, Web traffic~\cite{dataset:yahoo--webscopes5, xu2018unsupervised}, servers of Internet companies~\cite{dataset:aiops-kpi--ren2019time, dataset:omni-peidan--su2019robust}, spacecraft telemetry~\cite{dataset:nasa--hundman2018detecting}, and also medicine or robotics~\cite{dataset:illusion--wu2021current, dataset:numenta--ahmad2017unsupervised}. 
Due to the nature of the series, each anomaly (referred as an \emph{event} in the context of time series) can be a point in time (point-based anomaly) or occupy a range of consecutive samples (range-based anomaly).
The detection is performed in a supervised or in a unsupervised way, but the resulting performance of the algorithm is generally always assessed against ground truth labels that have been previously collected (either in controlled environments or labeled by experts in the field). 
This assessment is realized with \emph{evaluation metrics} taking as input both the ground truth and the predicted labels, and outputting one or multiple scores. 
The most common metrics for anomaly detection are the classical precision and recall, computed by comparing the predicted and the ground truth outputs for each sample.
In the usual terminology, the positive samples refer to the samples that are predicted as positive, and are partitioned into the true positives (TP, positive samples that are also anomalous in the ground truth) and false positives (FP). Likewise, the samples predicted as negative are partitioned into false negatives (FN) and true negatives (TN). 
The \emph{precision} measures the proportion $\text{TP}/(\text{TP}+\text{FP})$ of positive predicted samples that are correct, whereas the \emph{recall} measures the proportion $\text{TP}/(\text{TP}+\text{FN})$ of positive ground truth samples that have been retrieved.
Those classical metrics are convenient for the tasks that regard each sample separately, however this does not hold for time series datasets, where the time component is intrinsically continuous. 
Researchers developing detection algorithms have realized this challenge during the evaluation process and have come up with metrics fitting their specific use-case: range precision/recall~\cite{tatbul2018precision} for evaluating the Greenhouse algorithm~\cite{lee2018greenhouse}, time-aware precision/recall~\cite{hwang2019time} for evaluating the HAI dataset~\cite{dataset:water-hai--shin2020hai}, Numenta benchmark~\cite{numenta2015evaluating} for evaluating the Numenta corpus~\cite{dataset:numenta--ahmad2017unsupervised}, etc. 

In this paper, we first provide a comprehensive picture of the limitations of the classical metrics, along with the different directions of research that have been explored to handle them.
Against this background, we introduce a new pair of precision/recall metrics named the \emph{affiliation metrics} -- that exhibit a series of important properties as they are  theoretically principled, parameter-free, robust against adversary predictions, retain a physical meaning (as they are connected to quantities expressed in time units), and are locally interpretable (allowing to troubleshoot detection at individual event level). Summarizing our main contributions:
\begin{itemize}
    \item We show that existing range-based metrics for anomaly detection are easily gamed by adversary predictions: this complements dataset flaws outlined in~\cite{dataset:illusion--wu2021current}, and concur in creating an ``illusion of progress''.
    \item We introduce the affiliation metrics, an extension of the classical precision/recall for time series anomaly detection that is local, parameter-free, and applicable generically on both point and range-based anomalies.
    \item We produce closed-form expectations of the affiliation metrics in theoretical scenarios, proving their robustness against adversary strategies.
    \item We contrast the affiliation metrics to existing metrics in real datasets, and further show local interpretability at the event level, giving visual clues for algorithmic comparison.  
\end{itemize}

In the following sections, we detail the background for extending the metrics for time series (Sec.~\ref{sec2:background}) before introducing the affiliation metrics (Sec.~\ref{sec3:main_algo}). We then evaluate theoretical and practical properties of the proposed metrics (Sec.~\ref{sec4:evaluation}). We conclude by discussing the application scope of the affiliation metrics (Sec.~\ref{sec5:discussion} and Sec.~\ref{sec6:conclusions}).

\section{Background and motivation}\label{sec2:background}
We first introduce the limitations of the classical metrics for time series anomaly detection, which led to the design of new metrics that we briefly overview. We next illustrate the main limits of those proposed metrics, and sum up the main desirable goals for proper metrics definition, which motivated our work in the first place. 

\begin{figure}
  \centering
  \vspace{-0.2cm}
  \includegraphics[width=\linewidth]{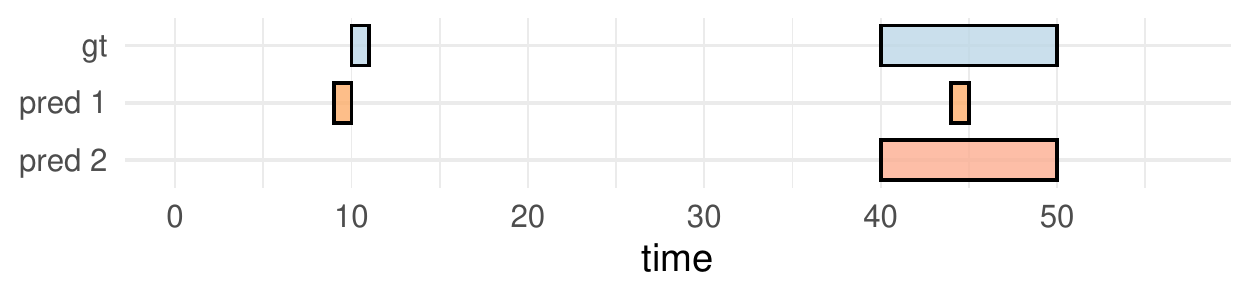}
  \vspace{-0.8cm} 
  \caption{The classical precision/recall of predictions 1 and 2 against ground truth are $0.50/0.09$ and $1.00/0.91$, illustrating resp. limitations (A), since each ground truth event is approximately detected but the scores are low, and (B), since only a single event is correctly detected but the scores are high.}\label{fig:revision_examplary_section2}
  \vspace{-0.2cm}
\end{figure}

\paragraph{\bf Limitations of the classical metrics for time series tasks}
As summarized in Tab.~\ref{tab:classical_limitations} and illustrated in Fig.~\ref{fig:revision_examplary_section2}, two main limitations have been observed in the literature for dealing with time series tasks (e.g., point or range anomaly detection, segmentation, or change point detection). 
The (A)~\emph{unawareness of temporal adjacency} prevents the metric from valuing the proximity between the samples. For instance, a prediction closely located in time to a ground truth label is adding both a FP and a FN samples instead of being considered as a TP, even for a one sample miss~\cite{gensler2014novel, scharwachter2020statistical, numenta2015evaluating, gharghabi2017matrix}.
Similarly, the predictions located closely after the end of a ground truth event (dubbed as ``ambiguous samples'' by Hwang et al.~\cite{hwang2019time}) are immediately penalized without any tolerance. 
The other aspect concerns the (B)~\emph{unawareness of the events durations}, which relates to evaluation of the individual samples without considering each event as a single unit.
Adverse consequences include the overrating of long events, in the sense that correctly detecting such event will be rewarded much more than correctly detecting another single-sample outlier~\cite{tatbul2018precision, hwang2019time}. 

\paragraph{\bf Recent time series evaluation metrics}
To cope with the above problems, numerous evaluation metrics have been recently introduced to better handle the time component, that  can be grouped into three main categories: (i) \emph{distance-based} metrics, (ii) \emph{window-based} metrics, and (iii) metrics specific to \emph{range-based} anomalies.

The first direction to handle near detection, i.e. limitation~(A), has been to employ direct measurement of the \emph{distances} between the elements from the two sets to derive a single score. This score is usually measured as a total deviation distance and is meant to be minimized. Since the Hausdorff distance is sensitive to the presence of any outlier, it is not suitable for evaluating the time series prediction tasks~\cite{fujita2013metrics, james2020novel}. 
Further work have therefore carried out modified Hausdorff distances~\cite{james2020novel} built on metrics introduced for computer vision tasks~\cite{dubuisson1994modified, deza2009encyclopedia}.

An orthogonal direction of research to handle  limitation~(A) has been to surround each ground truth event by a \emph{window}. Each window containing a predicted element is considered as a TP, relaxing the difficulty to obtain it compared to the classical matrix of confusion. The precision and recall are then deduced from this new counting. 
For point anomaly detection, Gensler and Sick~\cite{gensler2014novel} propose to count the predictions within a ground truth window as a TP, but only once for each window.
The same principle has been used in the context of change point detections~\cite{truong2020selective}.
The scoring of the ambiguous samples located after the anomalous point or range is also performed from ground truth windows: both in the Numenta Anomaly Benchmark (NAB) scoring~\cite{numenta2015evaluating} and in the time-aware precision/recall (TaP/TaR) metrics~\cite{hwang2019time}, predictions are mapped to a score based on a decaying sigmoid function.
Finally, Scharwächter and Müller~\cite{scharwachter2020statistical} compute the TP differently depending on the selected point of view, either from the predictions or from the ground truth.

To handle limitation~(B), \emph{range-based} metrics have recently been proposed~\cite{tatbul2018precision,hwang2019time,xu2018unsupervised,kim2021towards}. 
The range precision/recall (RP/RR)~\cite{tatbul2018precision} and the TaP/TaR~\cite{hwang2019time} are designed for anomaly detection events. 
The common idea consists in rewarding both the presence and the size of an overlap between the predicted and the ground truth events. For those metrics, each event is considered as a single unit irrespective of its length. 
Finally, the point adjust metrics~\cite{xu2018unsupervised} and an extension~\cite{kim2021towards} have been introduced to ease the scoring of the range-based events: the computation consists in sticking to the classical metrics, after initially extending each TP sample to the whole corresponding ground truth event. As~\cite{xu2018unsupervised, kim2021towards} however do not deal with limitation~(A) nor (B), this yield to a recent adaptation named F1-composite metric~\cite{garg2021evaluation}. 

\begin{table}[!t]
\caption{Limitations of the classical precision and recall for evaluating time series tasks as exposed in the literature.}\label{tab:classical_limitations}
\vspace{-0.1cm}
\begin{tabular}{@{}cccc@{}}
\toprule
& Limitation & Aspect & Mentions \\ \midrule
\multirow{2}{*}{(A)} & \multirow{2}{*}{\begin{tabular}[c]{@{}c@{}}Unawareness of the \\ temporal adjacency\end{tabular}} & \multirow{2}{*}{Inter-events} &  \multirow{2}{*}{\cite{gensler2014novel, scharwachter2020statistical, numenta2015evaluating, gharghabi2017matrix, hwang2019time}} \\ 
~ & ~ & ~ & ~ \\ 
\multirow{2}{*}{(B)} & \multirow{2}{*}{\begin{tabular}[c]{@{}c@{}}Unawareness of the \\ events durations\end{tabular}} & \multirow{2}{*}{Intra-event} & \multirow{2}{*}{\cite{tatbul2018precision, hwang2019time}} \\ 
~ & ~ & ~ & ~ \\ \bottomrule 
\end{tabular}
\end{table}

\begin{table}
  \caption{Characteristics of the recent metrics extending the classical precision/recall for time series anomaly detection.}\label{tab:new_limitations}
  \begin{tabular}{lcccc}
    \toprule
    (Class) Metric & \begin{tabular}[c]{@{}c@{}}P/R\\form?\end{tabular} & \begin{tabular}[c]{@{}c@{}}mainly\\for\end{tabular} & \begin{tabular}[c]{@{}c@{}}handles\\(A)~~~(B)\end{tabular} &    \begin{tabular}[c]{@{}c@{}}$\#$\\ param.\end{tabular}\\
    \midrule
    \rowcolor{grayshade}
      (i)\,\, distance~\cite{fujita2013metrics,james2020novel,gharghabi2017matrix} & \xmark & point & \cmark$\quad$\xmark & 0\\ 
    (ii) window~\cite{gensler2014novel, numenta2015evaluating, scharwachter2020statistical} & \cmark & point &
   \cmark$\quad$\xmark & 1\\
     \rowcolor{grayshade}
  $\quad$\,\,\,\,RP/RR~\cite{tatbul2018precision} & \cmark & range & \xmark$\quad$\cmark & 4\\
      \rowcolor{grayshade}
   (iii)  TaP/TaR~\cite{hwang2019time} & \cmark & range & \cmark$\quad$\cmark & 3\\
       \rowcolor{grayshade}
  $\quad$\,\,\,\,point adjust~\cite{xu2018unsupervised, kim2021towards} & \cmark & range & \xmark$\quad$\xmark & 0\\
     (---) affiliation (this work) & \cmark & both & \cmark$\quad$\cmark & 0 \\
    \bottomrule
  \end{tabular}
\end{table}

\paragraph{\bf Characteristics and limitations of the existing metrics}
The characteristics of the existing metrics are summarized in Tab.~\ref{tab:new_limitations}. Most of the metrics (except distance based) take the form of a precision and recall pair. Except TaP/TaR~\cite{hwang2019time}, no metric handles both limitations (A) and (B). Distance-based and windowed metrics are mostly limited to point anomalies and cannot handle~(B), while RP/RR and point adjust metrics are not able to handle limitation~(A), e.g. a one sample miss, since the overlap is empty in this case.  

Remarkably, with the exception of distance-based and point adjust metrics, \emph{one or multiple parameters} have to be selected to control several aspects of the scoring process, adding a number of parameters in the anomaly detection pipeline, likely leading to a lack of generality. 
While their purpose can be understood, fine-tuning or overspecializing the metric may be counterproductive for  real use-cases. Conversely while most of those parameters have default values, their setting is not always trivial. For instance, to handle (A)  both TaP/TaR and window-based metrics select a window size which is tuned for each dataset and drastically impacts the evaluation~\cite{scharwachter2020statistical, singh2017demystifying}. It even prevents the well-definition of the metrics when two windows overlap (i.e. leading to a precision possibly greater than one, as mentioned in~\cite{truong2020selective}, and occurring but not discussed for~\cite{gensler2014novel, numenta2015evaluating, hwang2019time}). 

\paragraph{\bf Interpretability and robustness against adversary algorithms or naive random predictions}
A core issue finally concerns the \emph{interpretability} of the recent range-based metrics. For instance RP/RR and TaP/TaR derive per event quantities, averaged over the whole dataset.
Individual quantities are however difficult to understand, as they combine multiple aspects: existence of an overlap, proportion of the overlap, relative positions, or ambiguous samples. Additionally, each predicted event is considered as an individual element, which is detrimental for the global meaning of the score, since the number of predicted events and their positions are not controlled: the resulting metric is not considering each region equally, that is any cluster of predictions in a specific region can impact globally the final score. This lack of locality allows the development of adversary algorithms reaching both high recall and high precision, as we show in Sec.~\ref{sec:adversary_predictions}. 

Another aspect related to interpretability concerns the lack of statistical properties constraining the behavior of the metrics. Overestimation of the scores has been shown for window metrics using Monte Carlo simulations~\cite{scharwachter2020statistical}, while  for the point adjust metrics it has been recently pointed out  that ``even a random anomaly score can easily turn into a state-of-the-art time series anomaly detection method''~\cite{kim2021towards} -- while in our proposal, random scores define the lower bound baseline as we formalize in 
Sec.~\ref{sec:properties} and demonstrate in Appendix~\ref{app:properties}.

\paragraph{\bf Design targets} 
To sum up, a convenient precision/recall pair for time series anomaly detection should include the following desirable targets, that have not been jointly addressed in prior art:
\begin{itemize}
    \item handle the limitations (A) and (B) introduced by the presence of time (only addressed by TaP/TaR);
    \item parameter-free definition (only available with  distance and point adjust metrics);
    \item expressiveness of the scores, in the sense that a slight improve of the predictions should result in a slight improve of the scores (only addressed by the distance metrics);
    \item local interpretability of the scores (not addressed so far);
    \item existence of statistical bounds of the scores (only addressed by simulation for window-based metrics).
\end{itemize}

\section{Local evaluation based on affiliation}\label{sec3:main_algo}

In this section, we introduce and describe the reasoning of the \emph{affiliation metrics} for evaluating anomalous events. Three concepts are developed. First, we define a directed average distance between sets, to measure how far the events are one from each other (Sec.~\ref{sec:average_distance_between_sets}). Then, each prediction is affiliated to the closest ground truth event, allowing a local perspective and maintaining the interpretability even for outlying predictions (Sec.~\ref{sec:aff_distance}). 
Finally, the observed temporal distances are locally converted into probabilities, by comparing them against random sampling, and are averaged into a precision/recall pair (Sec.~\ref{sec:aff_proba}). The practical aspects are detailed at the end of the section (Sec.~\ref{sec:practical_settings}).

We limitedly detail description for anomalous events consisting of ranges (particularization to the case of point anomalies is deferred to the Appendix~\ref{app:point_anomalies}).
An anomalous event is described by a continuous time interval $[t_{\text{start}}, t_{\text{stop}})$ with $t_{\text{stop}} > t_{\text{start}}$. Both prediction and ground truth are represented by a set of disjoint anomalous events, respectively noted $\text{pred}_1, \ldots \text{pred}_m$ and $\text{gt}_1, \ldots \text{gt}_n$. 

\subsection{Average distance between sets}\label{sec:average_distance_between_sets}

The distance from a point $x$ to a set $Y$ is defined, as usual, by: $\text{dist}(x,Y) := \min_{y \in Y} |x-y|$. For measuring the distance from a set $X$ to another, we consider the \emph{average directed distance} defined by:
$\text{dist}(X,Y) := \frac{1}{|X|} \int_{x \in X} \text{dist}(x, Y) dx.$
For the corner cases, we have $\text{dist}(X, \varnothing) = +\infty$ for nonempty $X$, and we keep $\text{dist}(\varnothing, Y)$ undefined for all $Y$.
This function is not a metric in the mathematical sense because it does not satisfy symmetry nor the triangle inequality, however is nonnegative and, for the case where $X,Y$ are sets of disjoint anomalous events, verifies $\text{dist}(X,Y) = 0 \Leftrightarrow (X \subset Y \text{ and } X \neq \varnothing)$. It has been introduced as a part of the modified Hausdorff distance~\cite{dubuisson1994modified} used in computer vision.

This directed distance has been selected for smoothness and interpretability reasons. First, contrary to the Hausdorff metric or to a simple threshold based on a window size, it satisfies smooth variation~\cite{dubuisson1994modified,james2020novel} since each sample contributes to the total score. 
Then, it has a clear interpretation as an average and retains a physical meaning as a time. Additionally, the distance is not converted into an undirected one -- such as taking the maximum over the two directions in the modified Hausdorff distance -- to prevent the dilution of the interpretation by an additional layer.
Indeed, our main idea in using this function is to relate the directed distance from prediction to ground truth events to a precision, and the one from the predicted events to the ground truth to a recall.

This idea is illustrated through the example in Fig.~\ref{fig:average_distance}. The prediction comprises three events whereas the ground truth is a single event.
From the predicted events to the ground truth (left of Fig.~\ref{fig:average_distance}), the directed distance is short, since $80\%$ of the prediction area matches with the ground truth while the remaining predicted event (accounting for $20\%$) is at a distance of $1\text{min}30\text{s}$ in average. In total, the directed distance is $18\text{s}$. 
This short distance is interpreted as a good precision of the predictions.
Improving the precision would be possible by removing the last predicted event and would lead to an average distance of zero corresponding to a perfect precision.
From the ground truth to the predicted events (right of Fig.~\ref{fig:average_distance}), the directed distance is computed in the other direction, giving a directed distance of $76.5\text{s}$. 
This distance is interpreted as a recall, currently expressed in time.
In that case, removing the last predicted event would not change the directed distance nor the recall.

\subsection{Local affiliation to the closest ground truth}\label{sec:aff_distance}

The time axis is partitioned by assigning each time $t$ to the closest ground truth event $\text{gt}_j$, $j \in \llbracket 1, n \rrbracket$. 
The resulting partition consists of $n$ intervals, and the $j$-th one $I_j$ is called the \emph{zone of affiliation} of the ground truth event $j$.
On each zone of affiliation, the ground truth and predictions belonging to it are obtained and the average directed distances described in Sec.~\ref{sec:average_distance_between_sets} are computed to retrieve the \emph{individual precision/recall distances}. 
Formally, the union of all predictions is noted $\text{pred} := \bigcup_{i=1}^m \text{pred}_i$ and the formulas are given, for $j \in \llbracket 1, n \rrbracket$, as follows:
\vspace{-0.05cm}
\begin{align}
    \text{D}_{\text{precision}_j}:=& \text{dist} \left(\text{pred} \cap I_j,~\text{gt}_j \right), \\
    \text{D}_{\text{recall}_j}:=& \text{dist} \left( \text{gt}_j,~\text{pred} \cap I_j \right).
\end{align}
The choice made by isolating each zone of affiliation from the ground truth perspective (for both precision/recall) is 
driven by the difference of reliability expected regarding the ground truth and the predictions.
For the ground truth, upstream control measures can be taken in place to assess its quality, for instance by ensuring that each labeled event corresponds to a separate anomaly that would need to be identified.
On the other hand, there should not be expectation for the shape of the predicted events, particularly on their distribution along the time line. As a consequence, taking the perspective of the ground truth is preventing the possibility for multiple short predictions stacked in a local area to bias globally the evaluation metric (cf.  Sec.~\ref{sec:adversary_predictions}).
Overall, each individual distance is related to a single ground truth and can be interpreted locally.
\begin{figure}[t]
  \centering
  \begin{subfigure}{1\linewidth}
    \includegraphics[width=\linewidth]{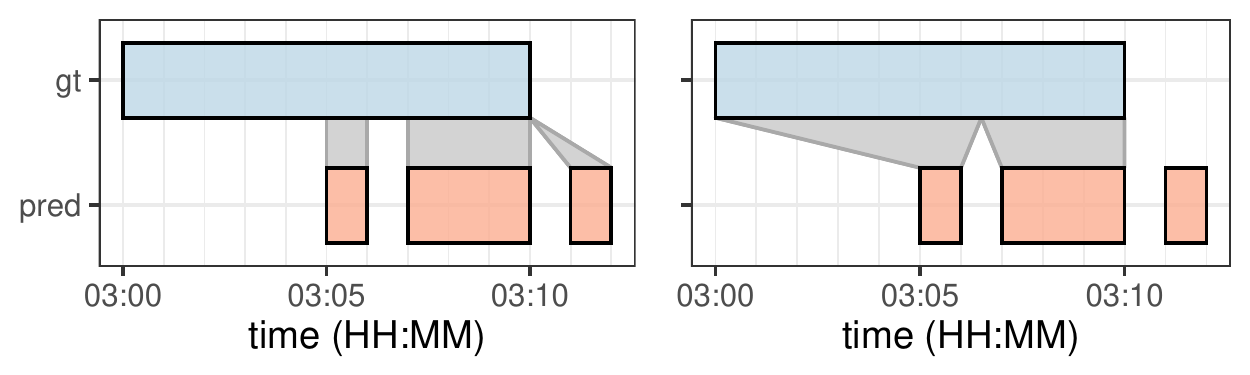}\vspace{-0.15cm}
    \caption{\emph{Average distance between sets}: example of the directed distance computed from predicted events to ground truth (left) and from ground truth to predicted events (right).}
    \label{fig:average_distance}
  \end{subfigure} \vspace{0.4cm} \\
  \begin{subfigure}{1\linewidth}
    \includegraphics[width=\linewidth]{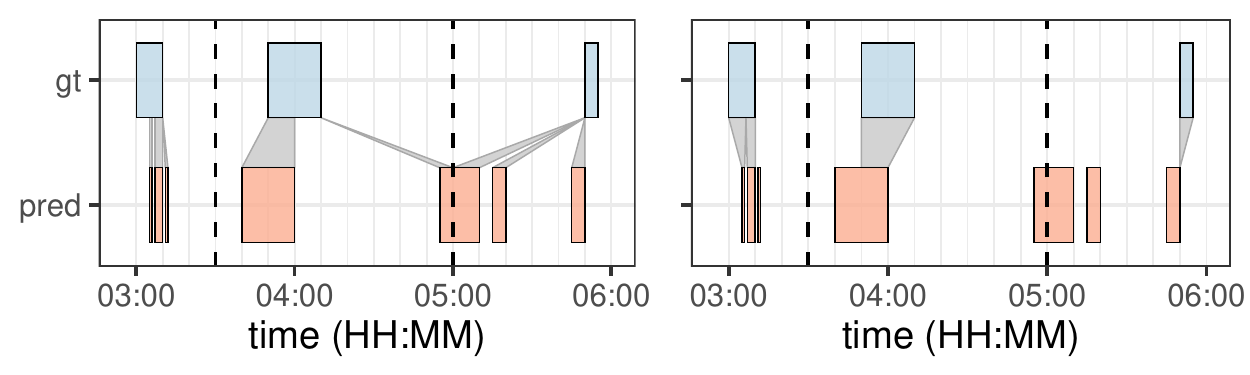}\vspace{-0.15cm}
    \caption{\emph{Local affiliation to the closest ground truth event}: example resulting in zones delimited by the dashed lines. The zones are similar for both directions: precision (left), and recall (right).}
    \label{fig:example_affiliation_metric}
  \end{subfigure} \vspace{0.4cm} \\
  \begin{subfigure}{1\linewidth}
    \hspace{0.65cm}$x \mapsto \bar{F}_{\text{precision}_j}( \text{dist}(x, \text{gt}_j))$\hspace{0.2cm}$y \mapsto \bar{F}_{y, \text{recall}_j}( \text{dist}(y, \text{pred} \cap I_j))$ \\ \vspace{-0.05cm}
    \includegraphics[width=\linewidth]{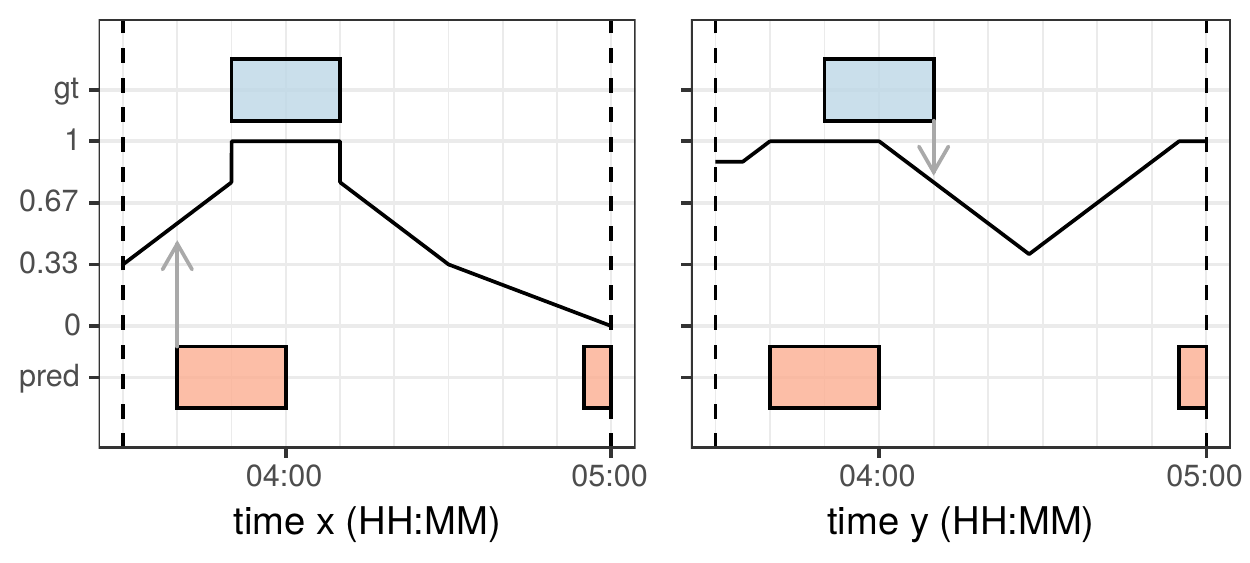}\vspace{-0.15cm}
    \caption{\emph{Comparison against random sampling}: example for converting each predicted sample to a precision score (left) and each ground truth sample to a recall score (right).}
    \label{fig:example_comparison_against_random}
  \end{subfigure}
  \caption{Illustration of the three steps for computing the affiliation metrics.}
  \vspace{-0.2cm}
\end{figure}

A practical derivation of the individual precision/recall distances is presented for the example shown in Fig.~\ref{fig:example_affiliation_metric}. First, the total timeline is cut into the zones of affiliation: $(-\infty, \text{3:30}), [\text{3:30}, \text{5:00}), [\text{5:00}, +\infty)$. Then the individual precision/recall distances are computed on each zone. For the first zone, it corresponds exactly to the case shown in Fig.~\ref{fig:average_distance}, giving $18\text{s}$~/~$76.5\text{s}$ for respectively the precision and the recall. The results are $11\text{min}30\text{s}$~/~$2\text{min}30\text{s}$ for the second zone and $31\text{min}15\text{s}$~/~$2\text{min}30\text{s}$ for the third one.

Individual distances are providing a summarized view of each ground truth event expressed in a meaningful unit (i.e., time), and can be used as they are by practitioners and domain experts. However, for comparing the different ground truth events of a dataset, it would be desirable to convert each individual value to the $[0,1]$ range, by assuming that each ground truth event is equally important (as opposed to the classical metrics that consider each sample as equally important). This normalization step is presented in Sec.~\ref{sec:aff_proba}.

\subsection{Comparison against random sampling}\label{sec:aff_proba}
The final normalization step consists in replacing each distance measured at the sample level with a probability $[0,1]$, by comparing this distance against the random sampling of a prediction.
We here introduce the main concept of this step, but we point out that the survival functions and the integrals involved in the computations have closed-form expressions, that are detailed in Appendix~\ref{app:closed_forms}. 

\paragraph{\bf Individual precision probability} 
As previously expressed, each affiliation zone is considered separately. On each zone, the ground truth $\text{gt}_j$ is fixed and a random prediction is made: this prediction $X$ is a random variable corresponding to a single point in time uniformly sampled within the affiliation zone. For the precision, the distance from $X$ to the ground truth $\text{gt}_j$ is a random variable with a cumulative distribution function $F_{\text{precision}_j}$ (that only depends on four time elements: the current zone and ground truth intervals).
The value of an observed distance $d \geq 0$ against this random prediction is defined by the survival function:
\begin{align}\label{eq:survival_precision}
    \bar{F}_{\text{precision}_j}(d) := 1 - F_{\text{precision}_j}(d-).
\end{align}
Applying this function on each predicted sample, we derive the \emph{individual precision probability} (see also Appendix \ref{appx:precision}) as follows:
\begin{align}\label{eq:individual_precision_probability}
    P_{\text{precision}_j} := \frac{1}{|\text{pred} \cap I_j|} \int_{x \in \text{pred} \cap I_j} \bar{F}_{\text{precision}_j}( \text{dist}(x, \text{gt}_j)) dx.
\end{align}

\paragraph{\bf Individual recall probability} For the recall, the distance depends on the considered ground truth sample $y \in \text{gt}_j$.
Knowing it, along with the affiliation zone interval, the distance from $y$ to $X$ is a random variable with a distribution $F_{y, \text{recall}_j}$.
As before, the value of an observed distance is defined by the survival function, and by applying this function on each ground truth sample, we derive the \emph{individual recall probability} as follows (see also Appendix \ref{appx:recall}):
\begin{align}\label{eq:individual_recall_probability}
    P_{\text{recall}_j} := \frac{1}{|\text{gt}_j|} \int_{y \in \text{gt}_j} \bar{F}_{y, \text{recall}_j}( \text{dist}(y, \text{pred} \cap I_j)) dy.
\end{align}

\paragraph{\bf Illustrative example} The computation for the second affiliation zone of Fig.~\ref{fig:example_affiliation_metric} is detailed with the help of Fig.~\ref{fig:example_comparison_against_random}.
For the precision, the function $x \mapsto \bar{F}_{\text{precision}_j}( \text{dist}(x, \text{gt}_j))$ is deduced from the survival distribution (left of Fig.~\ref{fig:example_comparison_against_random}). We observe that any predicted sample located inside the ground truth area would get a value of $1$, while this value decreases to $0$ as the distance to the ground truth event increases. For instance, the value of the predicted time $\text{3:40}$ is $0.556$ (illustrated by a gray arrow). Taking the average over all the predicted elements, we obtain an individual precision probability of $0.672$. 
For the recall, the function $y \mapsto \bar{F}_{y, \text{recall}_j}( \text{dist}(y, \text{pred} \cap I_j))$ is computed for each ground truth sample $y$ (right of  Fig.~\ref{fig:example_comparison_against_random}). 
Any ground truth element inside the predicted area has a value of $1$, that decreases as the distance to the predicted elements increases.
 
In the case of $y = \text{04:10}$, the evaluation gives a value of $0.778$ (illustrated by a gray arrow). 
Taking the average over all the ground truth elements, we obtain an individual recall probability of $0.944$.

\paragraph{\bf Interpretation and understanding} 
The baseline for the random prediction has been selected with the lightest possible a priori on the predicted samples, namely by selecting a single predicted sample randomly in each affiliation zone.
In particular, no information about the shape, the distribution, or the number of the predicted events is used.

Some observations can be directly made from the construction. First, we note that a precision of $1$ is equivalent to $\text{pred} \subset \text{gt}$, while a recall of $1$ corresponds to $\text{gt} \subset \text{pred}$, which is in alignment with the classical metrics.
Then, a precision or a recall lower than $0.5$ (resp. around $0.5$) is interpreted as doing worse than (resp. as bad as) a random prediction, meaning that the predictions made have not been able to provide information about the location of the event. 
The exact formulation of those interpretations constraining the behavior of the affiliation metrics are formalized through properties in Sec.~\ref{sec:properties}.

\paragraph{\bf Averaging of the individual precision/recall probabilities}

The \emph{precision/recall} are defined as the mean over the defined individual probabilities. Following the corner cases about the average distance between sets, an individual precision probability is undefined only when there is not any predicted element affiliated to $\text{gt}_j$. We let $S := \lbrace j \in \llbracket 1, n \rrbracket~;~\text{pred} \cap I_j \neq \varnothing \rbrace$, and obtain:


\begin{align}
  P_{\text{precision}} := \frac{1}{|S|} \sum_{j \in S} P_{\text{precision}_j}, \quad
  P_{\text{recall}} := \frac{1}{n} \sum_{j=1}^n P_{\text{recall}_j}.  
\end{align}

\subsection{Practical settings}\label{sec:practical_settings}

In real settings, the affiliation metrics can be applied on evenly or unevenly-spaced time series.
The ground truth and the predictions have the form of a binary vector of same length $N$, where the anomalous samples are indicated by $1$ and the normal ones by $0$, each index $i$ corresponding to a timestamp $t(i)$, as depicted in Tab.~\ref{tab:real_settings}.

\begin{table}[h!]
\caption{Illustration of the practical input shape for calculating the affiliation metrics, here corresponding to Fig.~\ref{fig:average_distance}}
\label{tab:real_settings}
\begin{tabular}{@{}cccccccccccccc@{}}
\toprule
index $i$ & 1 & 2 & 3 & 4 & 5 & 6 & 7 & 8 \\ \midrule
gt & 1 & 1 & 1 & 1 & 1 & 0 & 0 & 0 \\
pred & 0 & 0 & 1 & 0 & 1 & 0 & 1 & 0 \\
$t(i)$ & 3:00 & 3:02 & 3:05 & 3:06 & 3:07 & 3:10 & 3:11 & 3:12 \\ \bottomrule
\end{tabular}
\end{table}

A consistent way to convert those indexes into range-based events is to match any positive index $i$ to the corresponding interval $[t(i), t(i+1))$. The last timestamp $t(N+1)$ is clear for evenly spaced measures, otherwise it needs to be selected. In the example of Tab.~\ref{tab:real_settings}, it gives the events $\text{pred}_1 = [\text{3:05}, \text{3:06})$, $\text{pred}_2 = [\text{3:07}, \text{3:10})$, $\text{pred}_3 = [\text{3:11}, \text{3:12})$ and $\text{gt}_1 = [\text{3:00}, \text{3:10})$. It corresponds to the example of Fig.~\ref{fig:average_distance}.
Given those intervals, the individual precision/recall distances are available. With the additional knowledge of the total range $[t(1), t(N+1)]$, the precision/recall probabilities can be computed. Moreover, due
to the closed-form expressions developed in  Appendix~\ref{app:closed_forms},
their implementation is computationally efficient. Further practical considerations for reproducibility, including pointers to code, etc. are discussed in Appendix~\ref{app:code}. 

\section{Evaluation and properties}\label{sec4:evaluation}

The evaluation of the affiliated metrics is performed against the range-based anomaly detection metrics RP/RR and TaP/TaR, along with the classical sample-based metrics on a set of algorithms and datasets (Sec.~\ref{sec:settings}).
We first show that adversary predictions easily fool range metrics, whereas affiliated metrics are local and robust (Sec.~\ref{sec:adversary_predictions}). 
Second, we show that this local construction 
further allows a detailed per-event interpretation and comparison of the results given by the anomaly detection algorithms (Sec.~\ref{sec:interpretation}). Finally, we formalize (Sec.~\ref{sec:properties}) and prove (Appendix~\ref{app:properties})  theoretical properties of the affiliated metrics in typical prediction scenarios.

\subsection{Evaluation settings}\label{sec:settings}
\paragraph{\bf Benchmark anomaly detection algorithms}
As our primary aim is to contrast metrics for algorithmic evaluation, we rely on the anomaly detection algorithms that were selected by the previous authors~\cite{tatbul2018precision, hwang2019time}. For the datasets Machine-Temp, NYC-Taxi, and Twitter-AAPL, the predicted events are deduced from three algorithms: Greenhouse~\cite{lee2018greenhouse}, LSTM-AD~\cite{malhotra2015long}, both based on neural networks, and one implemented in the Luminol library based on time series bitmaps~\cite{wei2005assumption, luminol} (Luminol TSB). For the SWaT dataset, two unsupervised algorithms are selected (iForest~\cite{liu2008isolation} and OCSVM~\cite{scholkopf2001estimating}) along with a neural network approach (seq2seq~\cite{kim2019anomaly}).

\paragraph{\bf Datasets} Similarly, we select the four time series datasets used in~\cite{tatbul2018precision, hwang2019time} for the comparison.
Three of them are those used by authors of the RP/RR metrics~\cite{tatbul2018precision}, and taken from the publicly available Numenta Anomaly Benchmark Data Corpus~\cite{dataset:numenta--ahmad2017unsupervised}: \emph{Machine-Temp} (temperature sensor data of an industrial machine), \emph{NYC-Taxi} (number of New York City taxi passengers), and \emph{Twitter-AAPL} (collection of Twitter mentions of the ticker symbol AAPL). The other dataset 
has been used by authors of the TaP/TaR metrics~\cite{hwang2019time}, and is a secure water treatment testbed known as \emph{SWaT}~\cite{dataset:swat--goh2016dataset}.
A short description of the number of samples, percentage of anomalies, and number of anomalous events is available in Tab.~\ref{tab:datasets_description}.

\subsection{Adversary predictions}\label{sec:adversary_predictions}

\paragraph{\bf Adversary algorithm}
We design an algorithm to deceive metrics that consider each predicted event as a unit, by aggregating numerous predictions within a local region in order to impact the score globally.
Following the definition coined and discussed by Wu and Keogh~\cite{dataset:illusion--wu2021current}, an anomalous event is said \emph{trivial} if it can be identified with a single elementary line of code (e.g., a threshold)\footnote{Note that the threshold can be on the original data or from a new time series derived from it using basic primitive operations, such as a moving average of the series.}.
The function applied to derive a trivial event for the selected datasets is reported in Tab.~\ref{tab:datasets_description}. Regarding SWaT, since we do not have direct access  to the raw data, we simply consider one of the events, for instance the first one containing 940 samples, as trivial. 

Knowing any such trivial event, the adversary predictions are defined in two steps: (i) \emph{within the trivial event}, set the maximum possible number of predicted events by alternating positive and negative samples. (ii) \emph{outside the trivial event}, label all the samples as positive. We denote this  methodology for producing the predicted events as \emph{adversary algorithm}\footnote{We stress that similar adversary strategies can be defined for other metrics non included in the evaluation, such as  point adjust F1-composite metric~\cite{garg2021evaluation}. The adversary in this case would (i) keep the trivial event of duration $T$ and (ii) add $N$  additional predictions of length $\frac{T}{10N}$ regularly spaced over the rest of the interval, so that the precision is at least $T/(T+N(T/(10N))) = 1/(1+1/10) \approx 0.91$, while the recall is $1$ for sufficiently large $N$.}.
 
\begin{figure}
  \centering
  \includegraphics[width=\linewidth]{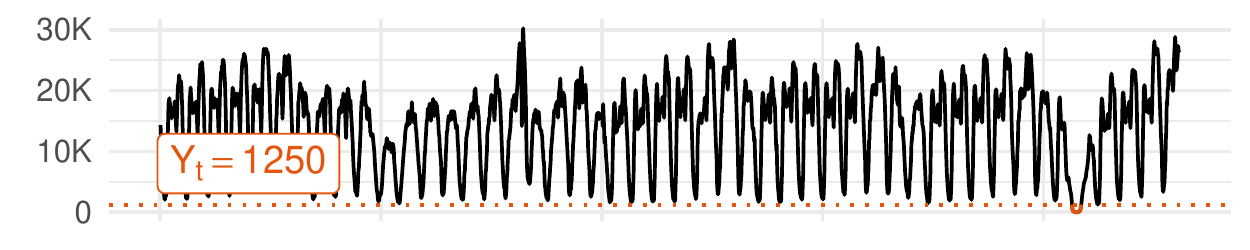}
  \includegraphics[width=\linewidth]{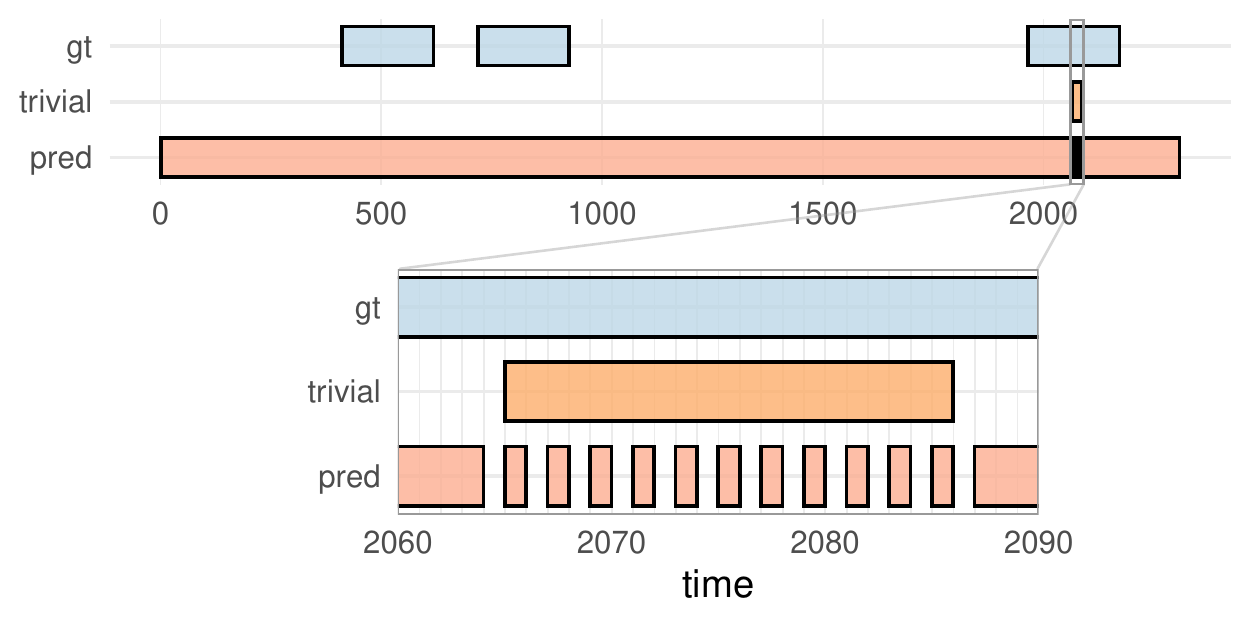}
  \vspace*{-0.8cm}
  \caption{Construction of the adversary predictions for the NYC-Taxi dataset, with a zoom on the interval $t \in [2060, 2090)$ containing the trivial event $Y_t < 1250$. The predicted events (bottom row) are calculated using the adversary algorithm given a trivial event (middle row). The predictions are evaluated against the ground truth (top row).}\label{fig:illustration_of_the_trivial_algorithm}
  \vspace{-0.2cm}
\end{figure}

The construction of the adversary is shown for the NYC-Taxi dataset  in  Fig.~\ref{fig:illustration_of_the_trivial_algorithm}. First, given the raw values of the series, a trivial event is identified (it does not need to cover entirely the true event). Then, the adversary algorithm is applied to produce the predictions, resulting in thirteen predicted events.

Trivial and adversary predictions are reported along with the other anomaly detection algorithms in Tab.~\ref{tab:metrics_trivial}.

\begin{table}
  \caption{Description of the datasets and of the corresponding base trivial events for the adversary algorithm.}
  \label{tab:datasets_description}
  \begin{tabular}{ccccc}
    \toprule
    & \begin{tabular}[c]{@{}c@{}}Machine\\ Temp\end{tabular} & \begin{tabular}[c]{@{}c@{}}NYC\\ Taxi\end{tabular} & \begin{tabular}[c]{@{}c@{}}Twitter\\ AAPL\end{tabular} & SWaT\\
    \midrule
    samples & 17682 & 2307 & 11889 & 449919\\
     anom. & 6\% & 27\% & 7\% & 12\%\\ 
    \# events & 2 & 3 & 2 & 35\\
    trivial & $Y_t < 40$ & $Y_t < 1250$ & $Y_t > 12000$ & First event \\
  \bottomrule
\end{tabular}
\end{table}

\begin{table*}
\caption{Comparison of the metrics using datasets and algorithm predictions selected by~\cite{tatbul2018precision,hwang2019time}, against a trivial prediction and an adversary algorithm. Each cell shows the precision/recall/F1-score w.r.t. a certain metric, algorithm, and dataset. The algorithm reaching the best F1-score is shown in bold for each metric and dataset.}\label{tab:metrics_trivial}
\begin{tabular}{@{}ccccc@{}}
\toprule
 \multicolumn{1}{c}{\multirow{2}{*}{\bf Metric}} &  \multicolumn{1}{c}{\multirow{2}{*}{\bf Algorithm}} & \multicolumn{3}{c}{\bf Dataset} \\ 
 &  & {\bf Machine-Temp} &  {\bf NYC-Taxi} & {\bf Twitter-AAPL} \\ \midrule
 & \advcell{trivial} & \advcell{\textbf{1.00/0.34/0.50}} & \advcell{1.00/0.03/0.07} & \advcell{\textbf{1.00/0.13/0.23}} \\
Classical & \advcell{adversary} & \advcell{0.05/0.83/0.10} & \advcell{\textbf{0.27/0.98/0.42}} & \advcell{0.06/0.93/0.12} \\
(sample-based & Greenhouse & 0.33/0.42/0.37 & 0.23/0.43/0.30 & 0.50/0.06/0.11 \\
precision/recall) & LSTM-AD & 0.06/1.00/0.12 & 0.24/0.49/0.32 & 0.24/0.13/0.17 \\
 & Luminol TSB & 0.10/0.04/0.06 & 0.15/0.02/0.04 & 0.37/0.07/0.11 \\ \midrule
 & \advcell{trivial} & \advcell{1.00/0.46/0.63} & \advcell{1.00/0.18/0.31} & \advcell{1.00/0.31/0.48} \\
 & \advcell{adversary} & \advcell{\textbf{0.99/0.75/0.85}} & \advcell{\textbf{0.88/0.85/0.86}} & \advcell{\textbf{0.96/0.75/0.85}} \\
RP/RR & Greenhouse & 0.15/0.57/0.24 & 0.23/0.52/0.32 & 0.26/0.51/0.35 \\
 & LSTM-AD & 0.03/1.00/0.06 & 0.27/0.51/0.35 & 0.10/0.51/0.17 \\
 & Luminol TSB & 0.08/0.50/0.14 & 0.14/0.34/0.20 & 0.24/0.51/0.32 \\ \midrule
 & \advcell{trivial} & \advcell{1.00/0.42/0.59} & \advcell{1.00/0.02/0.03} & \advcell{1.00/0.06/0.12} \\
 & \advcell{adversary} & \advcell{\textbf{0.99/0.96/0.97}} & \advcell{\textbf{0.93/1.00/0.96}} & \advcell{\textbf{0.96/1.00/0.98}} \\
TaP/TaR & Greenhouse & 0.17/0.47/0.25 & 0.32/0.64/0.43 & 0.42/0.04/0.07 \\
 & LSTM-AD & 0.04/1.00/0.07 & 0.36/0.66/0.47 & 0.13/0.08/0.10 \\
 & Luminol TSB & 0.10/0.02/0.04 & 0.23/0.02/0.03 & 0.26/0.04/0.07 \\ \midrule
 & \advcell{trivial} & \advcell{1.00/0.50/0.66} & \advcell{1.00/0.30/0.46} & \advcell{1.00/0.49/0.66} \\
 & \advcell{adversary} & \advcell{0.49/1.00/0.66} & \advcell{\textbf{0.54/1.00/0.70}} & \advcell{0.50/1.00/0.67} \\
Affiliation & Greenhouse & \textbf{0.71/0.99/0.83} & 0.51/0.99/0.67 & \textbf{0.78/0.98/0.87} \\
 & LSTM-AD & 0.50/1.00/0.67 & 0.51/1.00/0.67 & 0.66/0.99/0.79 \\
 & Luminol TSB & 0.54/0.99/0.70 & 0.38/0.79/0.51 & 0.73/0.98/0.83 \\ \bottomrule
\end{tabular}
\quad
\begin{tabular}{@{}cccc@{}}
\toprule
 \multicolumn{1}{c}{\multirow{2}{*}{\bf Algorithm}}
  & {\bf Dataset} \\  
  & {\bf SWaT} \\ \midrule
 \advcell{trivial} & \advcell{1.00/0.02/0.03} \\
 \advcell{adversary} & \advcell{0.12/0.99/0.21} \\
 iForest & \textbf{0.30/0.74/0.43} \\
 OCSVM & 0.17/0.85/0.28 \\ 
 seq2seq & 0.59/0.25/0.35 \\ \midrule
 \advcell{trivial} & \advcell{1.00/0.03/0.06} \\
 \advcell{adversary} & \advcell{\textbf{1.00/0.99/0.99}} \\
 iForest & 0.04/0.52/0.08 \\
 OCSVM & 0.14/0.61/0.23 \\ 
 seq2seq & 0.35/0.66/0.46 \\ \midrule
 \advcell{trivial} & \advcell{1.00/0.03/0.06} \\
 \advcell{adversary} & \advcell{\textbf{1.00/0.99/1.00}} \\
 iForest & 0.05/0.40/0.09 \\
 OCSVM & 0.17/0.55/0.26 \\ 
 seq2seq & 0.44/0.65/0.52 \\ \midrule
 \advcell{trivial} & \advcell{1.00/0.03/0.06} \\
 \advcell{adversary} & \advcell{0.53/1.00/0.69} \\
 iForest & 0.52/0.84/0.64 \\
 OCSVM & 0.65/0.70/0.68 \\ 
 seq2seq & \textbf{0.86/0.79/0.83} \\ \bottomrule
\end{tabular}
\end{table*}

\begin{table*}[t]
  \caption{Tabulated and visual per-event comparison between iForest and seq2seq on the SWaT dataset for the first six ground truth events, using the affiliation metrics. The affiliation zones are indicated with dashed  lines.}\label{tab:per_event}

\begin{tabular}{@{}cccccccc@{}}
\toprule
{\bf Algorithm  } & {\bf Mean of 35 events
} & {\bf  Ev. 1 } & {\bf Ev. 2 } & {\bf Ev. 3 } & {\bf Ev. 4 } & {\bf Ev. 5 } & {\bf Ev. 6 } \\ 
\midrule 
iForest & 0.52/0.84/0.64 & 0.37/0.53/0.44 & 
\textbf{1.00/0.91/0.95} & \textbf{0.76/0.99/0.86} & NaN/0/NaN & 0.38/0.60/0.46 & 0.09/0.21/0.12  \\
seq2seq &  \textbf{0.86/0.79/0.83} & \textbf{0.96/1.00/0.98} & 0.86/1.00/0.93 & 0.73/0.78/0.75 & \textbf{0.39/0.71/0.50} & \textbf{0.71/0.97/0.82} & \textbf{0.88/1.00/0.94}   \\ 
\bottomrule
\end{tabular}

\centering
  \includegraphics[width=\linewidth]{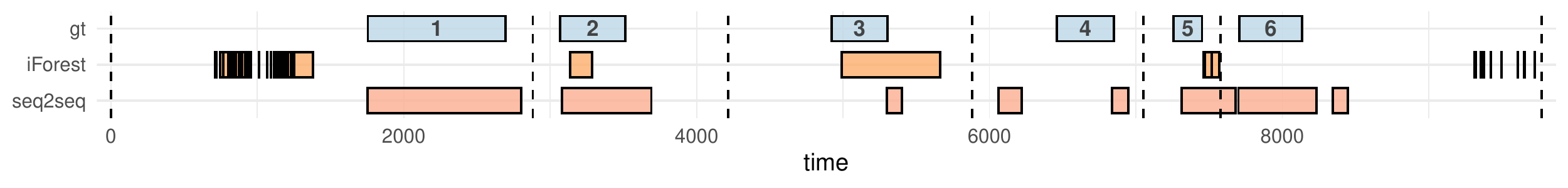}
  \vspace{-0.5cm}
\end{table*}

\paragraph{\bf Results for the RP/RR and TaP/TaR metrics}
The second and third rows of Tab.~\ref{tab:metrics_trivial} show that the adversary algorithm beats by far any other tested algorithms, although this adversary algorithm being unable to provide any informative content about the position of the anomalies, since almost all the samples are labelled positively. In this way, those algorithms are unable to provide reliable results in a conceivable situation.

The results of the evaluation for the adversary algorithm can be retrieved theoretically, for a dataset containing $n$ ground truth events and where the trivial event is cut into $k$ pieces. First, $n-1$ events are fully recalled, while the remaining one overlaps $50\%$ of the trivial event, giving a recall around $1 - \frac{1}{2n}$ for RP/RR and at least $1 - \frac{1}{4n}$ for TaP/TaR (using default parameters). 
Then, the precision is perfect for $k$ pieces and uncontrolled for the two remaining ones, giving a total precision of at least $\frac{k}{k+2}$.
Globally, both precision and recall are close to $1$ for $k$ and $n$ sufficiently large.

\paragraph{\bf Results for the classical and affiliation metrics} 
We observe in Tab.~\ref{tab:metrics_trivial} that the classical and the affiliation metrics are not sensitive to the adversary algorithm.

For the classical metrics, that operate at the \emph{sample level}, the trivial algorithm has a precision of one while the recall covers the proportion of correctly identified \emph{samples}. The adversary algorithm increases the recall but drastically impacts the precision, since most of the elements are now FP.

For the affiliation metrics, that operate at the \emph{event level}, the trivial algorithm has also precision of one while the recall covers the proportion of correctly identified \emph{events} (e.g. $1/35 \approx 0.03$ for the SWaT dataset). The adversary algorithm increases the recall but reduces the precision to $0.50$, meaning that the predictions are not better compared to a random prediction (by definition of the construction of the metric, as developed in Sec.~\ref{sec:properties}).

Regarding the other algorithms, we consider for instance Greenhouse and observe that it did not provide informative predictions for the NYC-Taxi (affiliation metrics with a precision around $0.50$), while performing better for both Machine-Temp and Twitter-AAPL.
For this latter, Greenhouse performs fewer predictions that do not cover the whole ground truth events but are overlapping or close to them, explaining the small classical recall (0.06) compared to the high affiliated recall (0.98).
This also holds for the SWaT dataset, for which we give a detailed event level interpretation in Sec.~\ref{sec:interpretation}.

\subsection{Event-level comparison}\label{sec:interpretation}
Since affiliation metrics have a local significance, each anomalous event can be analyzed individually from both precision and recall viewpoints, which is not possible with the other range-based metrics. 
An example of the results obtained using \emph{iForest} and \emph{seq2seq} for the 6 initial events of the $35$ overall anomalous events of the SWaT dataset  is shown in Tab.~\ref{tab:per_event}. The illustration intentionally mimics the one reported in~\cite{hwang2019time}, where  however the judgement of algorithmic performance at individual events level is left to the eye of the reader. In contrast, affiliation precisely quantifies in an unbiased and unequivocal manner the detection performance of each event,  offering a new light in the algorithmic evaluation through this new detailed view.

For events $1$ and $6$ of Tab.~\ref{tab:per_event}, iForest misses the ground truth events by far, while seq2seq predictions locate near and within the anomalous region. This translates into a poor precision/recall for iForest (worse than a random prediction for the precision, as bad as a random prediction for the recall) while for seq2seq the precision is good, with a perfect recall.
Event $2$ depicts the situation with a perfect precision (for iForest) or an almost perfect recall (for seq2seq), leading to similar performance since in both cases the main region has been correctly identified.
For event $3$, iForest has better identified the ground truth, even if both predictions are better than a random prediction. 
For event $4$, the prediction is either missing (for iForest) or poor (for seq2seq).
For event $5$, most the predictions made by iForest are close to the limit of the affiliation zone, impacting the overall precision, whereas seq2seq also captures most of the ground truth event inducing a boost in the precision and in the recall.

Overall, seq2seq gives better results for 21 events (13 for both precision and recall, and 8 only for the precision), equivocal results for 4 events (increase of the precision while the recall decreases, or the contrary) and a worse performance for 2 events. The remaining events are either completely undetected by iForest (for 2 events) or by seq2seq (for 6 events).

Globally, iForest, OCSVM, and adversary algorithms do not provide better results compared to a random guess. Locally, e.g. for events 2 and 3, the iForest algorithm is better (in terms of F1-score) compared to seq2seq. 
This behavior highlights the difference in the algorithm to deduce the anomalies (per sample for iForest and in-context for seq2seq) and can help for understanding the strengths and weaknesses of the algorithms and design better ones (for example for doing ensembles).

In an operational perspective, non-normalized distances can complement the understanding for each event by providing the individual time distances for precision/recall (optionally considering directionality, i.e., in case from a practical viewpoint an early detection is preferable to a late one).

\subsection{Theoretical properties}\label{sec:properties}

In addition with the practical results, we provide theoretical properties supporting the correct behavior of the affiliation metrics.
Since each affiliated zone is considered independently, we consider a single ground truth event $\text{gt}_j$ included in the affiliation zone $I_j$.
We let $p = |\text{gt}_j|/|I_j|$ the proportion taken by the ground truth event within its affiliation zone, which is also the proportion of positive samples.
As we consider an anomalous detection task, we are expecting rare events and $p \ll 1$ in most of the cases. 

In the following, we derive a closed-form of the metrics in three scenarios:  first (i) when the whole interval is predicted as anomalous, then (ii) for a random prediction within the affiliation zone, and finally (iii) for a single prediction located at specific locations on the interval. Details of the proofs are available in the Appendix~\ref{app:properties}.

\paragraph{\bf Predicting the whole interval as anomalous} In this case, the precision and recall are given by (cf. Appendix~\ref{appx:proof:whole} for proof):
\begin{align}\label{eq:property_whole}
    P_{\text{precision}} = \frac{1}{2} + \frac{p^2}{2},\quad P_{\text{recall}} = 1.
\end{align}

In the $p \ll 1$ regime, the precision is close to $1/2$ which corresponds to a poor detector (as poor as a random predictor).
For all values of $p$, this behavior can be put in parallel with the classical precision/recall, in which predicting all samples would give precision of $p$ and a recall of $1$.

\paragraph{\bf Expected precision and recall given a single random prediction} The expected precision and recall are given by:

\begin{align}
    P_{\text{precision}} = \frac{1}{2} + \frac{p^2}{2},\quad P_{\text{recall}} = \frac{1}{2}.
\end{align}

This property confirms that scores around $1/2$ corresponds to a random detector (cf. Appendix~\ref{appx:proof:single} for proof). In this case, the classical precision/recall would give a precision of $p$ and a recall close to $0$ assuming a large number of samples.

\paragraph{\bf Single prediction at a defined position}
We consider different single predictions located at four different positions on the affiliation zone: (a) the border of the affiliation zone, (b) the position halfway between the border and the ground truth event, (c) the first element of the event, and (d) the center of the event.
The latter case (d) corresponds to the best possible single-element prediction.
Since the position of the ground truth event also impacts the results, for simplification we assume it centered within the affiliation zone.

\begin{figure}[t]
  \centering
  \includegraphics[width=\linewidth]{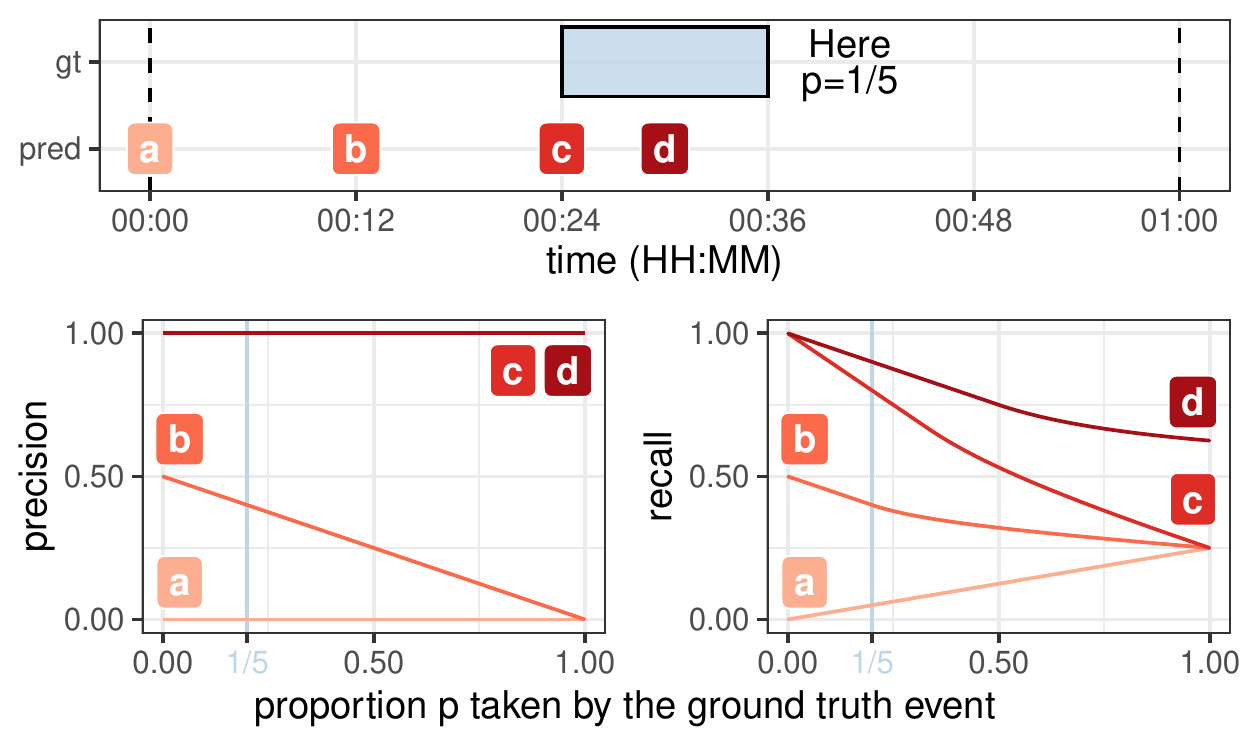}
  \vspace{-0.7cm}
  \caption{Given a ground truth centered in the affiliation zone and filling a proportion $p$, result of the affiliated metrics for a \emph{single-point} prediction located at the: (a)~border of the affiliation zone, (b)~position halfway between the border and the event, (c)~first element of the event, (d)~center of the event.}\label{fig:single_prediction}
\end{figure}

We defer a derivation of closed-form expression for (a)-(d) in Appendix~\ref{appx:proof:defined}, and report here an intuitive example. The scores as a function of $p$ are reported in Fig.~\ref{fig:single_prediction}, along with an illustration of the four positions for $p = 1/5$.
For $p \ll 1$, we observe that both precision and recall are $0$ for distant predictions (a), and increase until reaching $1$ for close predictions (c, d). In the other regimes, the precision is always $1$ for overlapping predictions (c, d) but the recall decreases as $p$ increases, representing the impossibility for a single prediction to reach a perfect recall of a large event.
For instance, the results for the best single-element prediction (d) are given by (with $x_{+} := \max \left( 0, x \right)$ the positive part):
\begin{align}\label{eq:centered_prediction_result}
    P_{\text{precision}} = 1,\quad P_{\text{recall}} = 1 - \frac{p}{2} + \frac{1}{2p} \left( p - \frac{1}{2} \right)_{+}^2.
\end{align}

\section{Discussion}\label{sec5:discussion}
We finally discuss the proposed affiliation metrics along complementary aspects.

\paragraph{\bf Shape of the ground truth labels}
The most important factor towards a correct evaluation is the ground truth labels themselves.
The intended purpose and a description of the labeling strategy are necessary requirements, but other factors impact the quality of the evaluation.
First, the actual output should represent an anomaly detection task, i.e. labeled events need to be rare.
Furthermore, we expect that each event corresponds to a single anomaly. For this purpose, the merge of fragmented events related to a single anomaly have been proposed~\cite{dataset:illusion--wu2021current}. 
However, the amplitude of the labeled zone may remain imprecise: even for controlled experiments~\cite{dataset:swat--goh2016dataset, dataset:water-hai--shin2020hai} for which the start date is known, the time at which the system returns to a steady state remains subjective. 
By design, the affiliation metrics are less sensitive to precise labeling compared to the previous range or window-based metrics.

\paragraph{\bf Expressiveness of the metrics} 
The affiliation metrics have been built to handle the two main limitations of the classical precision/recall. 
As for algorithmic comparison, we have shown  that they are expressive enough to perform a fair comparison without the need for introducing additional parameters.
The modification of the metrics to handle additional features (such as focusing on overlaps only, or giving more importance to the beginning of an event) is possible by modifying the survival functions -- but, at the same time, this is not encouraged.  The main reason is the additional layer of complexity needed to select those parameters, in the context of small volume of labeled events, would render evaluation arbitrary.

\paragraph{\bf Theoretical properties} 
The affiliation metrics focus on a single aspect of the evaluation: the assessment of the proximity between the predicted and the ground truth labels, which is the primary aspect to compare anomaly detection algorithms from a research point of view. Summarizations of the precision/recall as a single quantity that exist for the classical metrics (such as the F-score or the Average Precision) are straightforward to derive for the affiliation metrics. 
In order to properly characterize the behavior of such summarized metrics, further research is needed to gather theoretical bounds on the variance.

\paragraph{\bf Practical deployment} 
From a deployment perspective, complementary measures need to be considered, similarly to those introduced by Gensler and Sick~\cite{gensler2014novel}, such as the number of predicted events in each segmentation zone and the direction tendency of the predictions.
In this context, the trade-offs between those measures remain in the hand of the field expert. In order to assist decisions (e.g., algorithm selection and score thresholding), further development of an interactive visualization library leveraging the affiliation metrics would be beneficial.

\section{Conclusion}\label{sec6:conclusions}

For evaluating time series anomaly detection tasks, we proposed a precision/recall pair that handles the limitations encountered with the classical metrics.
Contrary to the existing metrics, it is generic (parameter-free, applicable on all datasets),
and local (each ground truth event is considered separately). In turn, locality makes it both 
expressive (possible to break down the final score into individual interpretable and visualizable bricks) and robust (e.g., not sensitive to adversary predictions).
Finally, its construction makes it both theoretically principled, as well as practically useful -- overall, we hope that the research community will find them a useful contribution  for the unbiased evaluation of time series anomaly detection tasks.

\printbibliography

\appendix

\section{Closed-form of the affiliation metrics}\label{app:closed_forms}

For the ground truth event $\text{gt}_j=:[a,b)$ located within the affiliation zone $I_j=:[A, B)$, we let respectively $m$ and $M$ the shortest and largest distance from the event to the borders of the zone:
$$m:= \min \left(a-A, B-b \right) \quad\text{ and }\quad M:= \max \left(a-A, B-b \right).$$

\subsection{Survival functions}

The survival function for the precision defined by Eq.~\ref{eq:survival_precision} is given by $\bar{F}_{\text{precision}_j}(0) = 1$ and, for $d \in (0, M]$: 
\begin{align}\label{eq:survival_precision_appendix}
\bar{F}_{\text{precision}_j}(d) = 1 - \frac{|\text{gt}_j| + \min(d, m) + d}{|I_j|}.
\end{align}

The explanation is as follows: taking uniformly at random an element within the affiliation zone, the probability to obtain a distance of zero is $|\text{gt}_j|/|I_j|$, while outside the decrease of the survival function is in $2d$ on $(0, m]$ and in $d$ on $(m, M]$, because two elements exist for each distance before $m$, and only one after.
For a ground truth sample $y \in \text{gt}_j$, we let:
$$m_y:= \min \left(y-A, B-y \right) \quad\text{ and }\quad M_y:= \max \left(y-A, B-y \right).$$
The survival function for the recall is given, for $d \in [0, M_y]$, by:
\begin{align}\label{eq:survival_recall_appendix}
\bar{F}_{y, \text{recall}_j}(d) = 1 - \frac{\min(d,m_y) + d}{|I_j|}.
\end{align}

\subsection{Closed-form of the integrals}

The closed-form of the integrals is straightforward (integral of piecewise linear functions) but include many cases due to the presence of the $\min$ and $\max$ functions. The expressions are derived in the following paragraphs. All the cases have been considered in the Python implementation~\cite{code_package}.

\subsubsection{\bf Integral over the samples for the precision}\label{appx:precision}
For the precision, each predicted interval located within the affiliation zone is cut into three pieces (possibly empty), corresponding to the portion before, within, and after the ground truth event.
We consider the case of a single predicted interval $\text{pred}$ located after the ground truth event (thus not intersecting it). 
The distance from the interval to the ground truth event goes from $d_{\text{min}} = \text{pred}_{\text{start}} - b$ to $d_{\text{max}} = \text{pred}_{\text{stop}} - b$. We also consider either the case (a) $d_\text{max} \leq m$ or (b) $m \leq d_\text{min}$. For the remaining case $m \in (d_\text{min}, d_\text{max})$, the predicted interval is cut again into two smaller pieces, one verifying (a) and the other (b). The integral over the samples is given by:
\begin{align}
    \mathcal{A} := \int_{x \in \text{pred}} \bar{F}_{\text{precision}_j}( \text{dist}(x, \text{gt}_j)) dx = \int_{d_{\text{min}}}^{d_{\text{max}}} \bar{F}_{\text{precision}_j}(z) dz.
\end{align}

We replace the survival function using (\ref{eq:survival_precision_appendix}). The linear part gives $\int_{d_{\text{min}}}^{d_{\text{max}}} z dz = (d_{\text{max}} - d_{\text{min}}) (d_{\text{max}} + d_{\text{min}})/2 = |\text{pred}|d_{\text{center}}$, with $d_{\text{center}}$ the distance reached for the point at the middle of the predicted interval. 
For the two cases (a) and (b), we deduce the same expression: 
\begin{align}\label{eq:closed_form_precision}
   \frac{\mathcal{A}}{|\text{pred}|} =  1 - \frac{|\text{gt}_j| +  \min(d_{\text{center}}, m)  + d_{\text{center}}}{|I_j|}.
\end{align}

\subsubsection{\bf Integral over the samples for the recall}\label{appx:recall} For the recall, we cut the ground truth interval into a finite partition such that each piece $\text{gt}_{j,k}$ is either fully included in the predictions (and the integral is immediate) or has a unique closest prediction $t_{\text{pivot}, k} \in I_j$ located at the border or outside the piece. Furthermore, all elements of a piece should be closer either to the closest prediction or to the border.
We consider the case of a closest prediction located after the ground truth event, so that the distance from an element $y \in \text{gt}_{j,k}$ to the predictions is: $t_{\text{pivot}, k} - y$.

The integral over the samples is given by:
\begin{align}
    \mathcal{B}:=& \int_{y \in \text{gt}_{j,k}} \bar{F}_{y, \text{recall}_j}( \text{dist}(y, \text{pred} \cap I_j)) dy \\
    =& \int_{y \in \text{gt}_{j,k}} \bar{F}_{y, \text{recall}_j}(t_{\text{pivot}, k} - y) dy.
\end{align}

We replace the survival function using Eq.~\ref{eq:survival_recall_appendix}. We define $y_{\text{center}}$ the middle point of $\text{gt}_{j,k}$, $d_{\text{pivot}} = |t_{\text{pivot}, k}  - y_{\text{center}}|$ its distance to the pivot prediction, and $m_{\text{center}} = \min(y_{\text{center}} - A, B - y_{\text{center}})$ its closest distance to the border. 
Using those notations, we obtain on the one side the linear part:
\begin{align}
    \int_{y \in \text{gt}_{j,k}} (t_{\text{pivot}, k} - y) dy = |\text{gt}_{j,k}| d_{\text{pivot}}
\end{align}
and on the other side:
\begin{align}
\int_{y \in \text{gt}_{j,k}} \min (t_{\text{pivot}, k} - y, m_y) dy = |\text{gt}_{j,k}| \min (d_{\text{pivot}}, m_{\text{center}} ).
\end{align}
Combining those elements, we arrive at the following expression:

\begin{align}\label{eq:closed_form_recall}
   \frac{\mathcal{B}}{|\text{gt}_{j,k}|} =  1 - \frac{\min(d_{\text{pivot}}, m_{\text{center}})  + d_{\text{pivot}}}{|I_j|}.
\end{align}

\section{Particularization to point anomalies}\label{app:point_anomalies}

The case of point anomalies corresponds to express each anomaly at date $t$ as the limit of an event $[t, t+\varepsilon)$ when $\varepsilon \to 0$. It is therefore immediate to restate Eq.~\ref{eq:individual_precision_probability} as follows (with $\text{pred}$ corresponding here to the point predictions within the affiliation zone $I_j$):

\begin{align}
    P_{\text{precision}_j} = \frac{1}{ \# \text{pred}} \sum_{x \in \text{pred}} \bar{F}_{\text{precision}_j}( \text{dist}(x, \text{gt}_j))
\end{align}

and Eq.~\ref{eq:individual_recall_probability}, since $\text{gt}_j$ is now a single point, as:
\begin{align}
    P_{\text{recall}_j} = \bar{F}_{\text{gt}_j, \text{recall}_j}( \text{dist}(\text{gt}_j, \text{pred})).
\end{align}

The forms of the survival functions do not change (in Eq.~\ref{eq:survival_precision_appendix}, the term $|\text{gt}_j|$ is replaced by $0$).

\section{Proof of the properties}\label{app:properties}

The generic method for proving the properties is to find a cut of the intervals satisfying the conditions expressed in Appendix~\ref{app:closed_forms}, and to apply   (\ref{eq:closed_form_precision}) and~(\ref{eq:closed_form_recall}).

\subsection{Predicting the whole interval as anomalous}\label{appx:proof:whole}
The recall is $1$ because the ground truth interval is included in the predictions. For the precision, we suppose that $m = a-A$ (the other case is symmetric).
The predicted interval $[A,B)$ is cut into $[A,a) \cup [a,b) \cup [b,b+m) \cup [b+m, B)$. The Eq.~\ref{eq:closed_form_precision} is applied on each part, giving four areas: 
\begin{align*}
    &\mathcal{A}_{[A,a)} = \frac{(a-A)(B-b)}{B-A}~~,~~\mathcal{A}_{[a,b)} = b-a~~,~~\mathcal{A}_{[b,b+m)} = \mathcal{A}_{[A,a)}, \\
    &\mathcal{A}_{[b+m, B)} = \frac{1}{2} \frac{((B-b)-(a-A))^2}{B-A}.
\end{align*}
It leads to the following expression (corresponding to the Eq.~\ref{eq:property_whole}):
\begin{align}\label{eq:final_whole_interval_precision}
P_\text{precision} = \frac{\mathcal{A}_{[A,a)} + \mathcal{A}_{[A,a)} + \mathcal{A}_{[A,a)} + \mathcal{A}_{[A,a)}}{B-A} = \frac{1}{2} + \frac{1}{2} \left(\frac{b-a}{B-A}\right)^2.
\end{align}

\subsection{Expected precision and recall given a single random prediction}\label{appx:proof:single}
The expected recall is expressed as the average of (\ref{eq:individual_recall_probability}) over $t \in I_j$: 
\begin{align*}
    \mathbb{E} \left[ P_{\text{recall}} \right] &= \frac{1}{|\text{gt}_j|} \frac{1}{|I_j|} \int_{t \in I_j} \int_{y \in \text{gt}_j} \bar{F}_{y, \text{recall}_j}( \text{dist}(y, t)) dy dt \\
    &= \frac{1}{|\text{gt}_j|} \frac{1}{|I_j|} \int_{y} \int_{t} 1 - \frac{\min(|t-y|, m_y) + |t-y|}{|I_j|} dt dy
\end{align*}
\noindent where on the second line, we use the Fubini's theorem, which facilitates the computation of the inner integral for each fixed $y$:

\begin{align*}
R(y) :=& \int_{t} 1 - \frac{\min(|t-y|, m_y) + |t-y|}{|I_j|} dt \\
=& |I_j| - \frac{1}{|I_j|} \left[ \int \min(|t-y|, m_y) + |t-y| dt \right] \\
=& |I_j| - \frac{1}{|I_j|} \left[ 2 \int_{0}^{m_y} 2zdz + \int_{m_y}^{M_y} (m_y + z)dz \right].
\end{align*}

For the last line, we observe that for $t \in [y - m_{y}, y + m_{y}]$, the distance $|t-y|$ goes from $0$ to $m_y$ (and this happens two times, on the left and on the right), while for $t \in [m_y, M_y]$, the distance goes from $m_y$ to $M_y$.
After computing the integral, we end up with a quantity which does not depend on $y$: 
\begin{align}
R(y) = |I_j| - \frac{1}{2|I_j|} (m_y + M_y)^2 = \frac{|I_j|}{2},
\end{align}
and finally:
    $\mathbb{E} \left[ P_{\text{recall}} \right] = 1/2.$ 
 For the expected precision, the computations are identical to those leading to Eq.~\ref{eq:final_whole_interval_precision}.
 
\subsection{Single prediction at a defined position}\label{appx:proof:defined}

To ease the understanding, we consider the case $|I_j|=[A,B)=[0,1)$. Since $\text{gt}_j = [a,b)$ is centered in the affiliation zone and of proportion $p$, we have: $[a,b) = [1/2-p/2, 1/2+p/2)$. We detail the case (d) of a prediction located at the position $1/2$ and corresponding to Eq.~\ref{eq:centered_prediction_result}.

If $p \leq 1/2$, the cut $[1/2-p/2, 1/2) \cup [1/2, 1/2+p/2)$ verifies the necessary conditions. 
By applying Eq.~\ref{eq:closed_form_recall} on each part, we obtain:
    $\mathcal{B}_{[1/2-p/2, 1/2)} = \mathcal{B}_{[1/2, 1/2+p/2)} = (p/2) (1 - p/2)$
so that $P_{\text{recall}} = 1 - p/2.$

If $p > 1/2$, the ground truth region is partitioned into four areas: $[1/2-p/2, 1/4) \cup [1/4, 1/2) \cup [1/2, 3/4) \cup [3/4, 1/2+p/2),$ from which we compute:
\begin{align*}
\mathcal{B}_{[1/4, 1/2)} = \mathcal{B}_{[1/2, 3/4)} =& (1/4) (3/4), \\
\mathcal{B}_{[1/2-p/2, 1/4)} =  \mathcal{B}_{[3/4, 1/2+p/2)} =& (p/2-1/4) (1/2), 
\end{align*}
so that $P_{\text{recall}} = 1/2 + 1/(8p)$. 
By combining those two cases, we end up with Eq.~\ref{eq:centered_prediction_result}.
Globally, the same method is applied for the positions (a), (b), (c), and we get the curves represented in Fig.~\ref{fig:single_prediction}:
\vspace{0.1cm}
\begin{align}\label{eq:single_prediction_a}
    \text{(a)}& \quad P_{\text{precision}} = 0,~ &&P_{\text{recall}} =&& \frac{p}{4}, \\
    \text{(b)}&  \quad P_{\text{precision}} = \frac{1}{2} - \frac{p}{2},~ &&P_{\text{recall}} =&& 
    \frac{1}{2} - \frac{p}{2} + \frac{25}{64p} \left(p - \frac{1}{5} \right)_{+}^2, \\ 
    \text{(c)}& \quad P_{\text{precision}} = 1,~ &&P_{\text{recall}} =&& 1 - p + \frac{16}{9p} \left(p - \frac{1}{3} \right)_{+}^2, \\ 
    \text{(d)}& \quad P_{\text{precision}} = 1,~ &&P_{\text{recall}} =&& 1 - \frac{p}{2} + \frac{1}{2p} \left( p - \frac{1}{2} \right)_{+}^2. 
\end{align}

\section{Reproducibility}\label{app:code}

The affiliation metrics have been implemented using the standard Python 3 library and is available at~\cite{code_package}.
The implementation leverages the closed-form highlighted in Appendix~\ref{app:closed_forms} and follows the process of Sec.~\ref{sec:practical_settings}: the binary inputs are converted into events before being separated into affiliation zones. On each segment, the metrics are computed and the output consists of the precision/recall scores as well as the individual distances and probabilities.

Reliability of the code has been checked through unit tests.
Additionally, the numerical results related to the affiliation metrics obtained in Sec.~\ref{sec4:evaluation} are directly reproducible by typing the following: \texttt{python -m unittest discover}.
On a Windows 10 machine using an Intel(R) Core(TM) i7-8650U processor running at 1.90 GHz with 16 GB of RAM, the whole tests took 8 seconds, including the computation of the affiliation metrics for the SWaT~\cite{dataset:swat--goh2016dataset} dataset containing 449919 samples and 35 ground truth events, against up to 472 predicted events. 

\end{document}